\documentclass[
]{ceurart}

\usepackage{listings}
\begin{document}

\copyrightyear{2025}
\copyrightclause{Copyright for this paper by its authors. Use permitted under Creative Commons License Attribution 4.0 International (CC BY 4.0). Originally published online by CEUR Workshop Proceedings (CEUR-WS.org, ISSN 1613-0073) within ISWC~2025 Companion Volume, available online: https://ceur-ws.org/Vol-4085/paper9.pdf}
\conference{Originally published online by CEUR Workshop Proceedings (CEUR-WS.org, ISSN 1613-0073) within ISWC~2025 Companion Volume. Available online: \url{https://ceur-ws.org/Vol-4085/} and \url{https://ceur-ws.org/Vol-4085/paper9.pdf}}



\title{Mitigating Undesired Conditions in Flexible Production with Product--Process--Resource Asset Knowledge Graphs}


\author[1]{Petr Novák}[%
orcid=0000-0003-1720-7334,
email=petr.novak@cvut.cz,
url=https://www.ciirc.cvut.cz/,
]
\cormark[1]
\fnmark[1]
\address[1]{Czech Institute of Informatics, Robotics and Cybernetics, Czech Technical University in Prague, Jugoslávských partyzánů 1580/3, Prague, CZ-16000, Czech Republic}

\author[2]{Stefan Biffl}[%
orcid=0000-0002-3413-7780,
email=stefan.biffl@tuwien.ac.at,
url=https://www.tuwien.ac.at/,
]
\fnmark[1]
\address[2]{TU Wien -- Institute of Information Systems Engineering, Faculty of Informatics, Favoritenstrasse 9-11, Vienna, A-1040, Austria}

\author[3]{Marek Obitko}[%
email=mobitko@ra.rockwell.com,
orcid=0000-0001-7639-3312,
url=https://www.rockwellautomation.com/en-cz.html,
]
\fnmark[1]
\address[3]{Rockwell Automation -- Advanced Technology, Argentinská 1610/4, Prague, CZ-17000, Czech Republic}

\author[1]{Petr Kadera}[%
orcid=0000-0002-1747-6473,
email=petr.kadera@cvut.cz,
url=https://www.ciirc.cvut.cz/
]
\fnmark[1]

\cortext[1]{Corresponding author.}
\fntext[1]{These authors contributed equally.}


\maketitle


\section{Introduction}

The evolution of manufacturing systems towards Industry~4.0, bringing flexibility of production processes, re-configurability of production system resources, and better use of industrial artificial intelligence, has fundamentally transformed the operational landscape of industrial production environments~\cite{Naeem2024}. 
Industrial cyber-physical production systems conforming to Industry~4.0 represent software-intensive architectures, where industrial workcells, robots, and other resources incorporate decisive roles in system functionality, encompassing robot control programs, as well as Manufacturing Execution Systems 
or Manufacturing Operations Management 
systems that orchestrate production at the system level~\cite{Novak2022PyMES}. 
This transformation from centralized to distributed system architectures introduces unprecedented levels of flexibility, enabling mass customization and adaptive manufacturing~\cite{Lunsch2022,Wally2021TASE}.
As a consequence, production engineering has to shift from a component viewpoint to a system viewpoint and to consider holistic principles of system engineering and analysis~\cite{Winkler2021Risk}. 
 
However, such an increased flexibility brought by Industry 4.0 or smart manufacturing presents significant challenges to traditional quality assurance and error diagnostic methodologies~\cite{Wehner2023Interactive}. 
Unlike conventional manufacturing systems that relied on repeatable production sequences executed on identical machines with standardized processes, modern cyber-physical production systems (CPPSs) exhibit dynamic behavior patterns that complicate the testing or commissioning processes, fault detection, and quality control on the system level. 
The inherent variability in flexible manufacturing makes traditional statistical process control methods inadequate for comprehensive quality management as products are frequently produced on different machines/resources with various production processes, making production processes hard to repeat and significantly more difficult to test and fine-tune.
As flexible use cases make it no longer efficient or even feasible to rely on traditional Ishikawa/fishbone diagrams~\cite{Winkler2021Risk}, industrial practitioners require more versatile, flexible, and dynamic models and approaches for operations support and quality assurance.


To cope with all these challenges and constraints, we designed a semantic model realizing the product--process--resource (PPR) asset knowledge graph (AKG). The proposed combined model fulfills the following requirements, substantiated by recommendations from industrial partner practitioners:
\begin{itemize}
    \item Req. 1 -- Model ability to represent flexible Industry~4.0 production of products with highly automated production resources, adopting principles of skill-based engineering;
    \item Req. 2 -- Model ability to represent undesired conditions relevant for production system engineering in terms of plausible causes and effects;
    \item Req. 3 -- Support for efficient interaction with human engineers and operators via LLMs/chatbots.
\end{itemize}

The PPR-AKG approach has been used for handing our laboratory prototype over to an industrial vendor in a project: Automated robotic disassembly of used electric vehicle batteries and their remanufacturing~\cite{Chigbu2024EVBattRemanufacturing} for stationary energy storages~\cite{Strakosova2024ETFAPoPAN}.
This use case involves highly variable input product and material states, affecting subsequent processes, and complex quality assessment procedures.


\section{PPR Asset Knowledge Graph (PPR-AKG) for Modeling Production Processes and Systems with Undesired Conditions and their Causes }

The proposed approach ``Product--Process--Resource Asset Knowledge Graphs'' (PPR-AKGs) extends a Product--Process--Resource (PPR) model, established in~\cite{Schleipen2009} and grounded in VDI~3682, ISA-95, and IEC~62264.
The PPR-AKG provides a foundational manufacturing system model~\cite{Biffl2021} by recognizing triadic relationship between products, production processes, and production resources for comprehensive system understanding. 
However, the traditional representations 
lack the identification of undesired and desired conditions, which are crucial for practical operation in industrial manufacturing. 
The proposed PPR-AKG model shall close this gap by extending the PPR model with undesired conditions and their causes. 
In addition, traditional PPR implementations lack the semantic richness necessary for expressing complex dependencies and causal relationships that characterize modern CPPS operations~\cite{Terkaj2022}. 
Therefore, the comprehensive model is represented as OWL (i.e., Web Ontology Language) ontology, benefiting from ontology features such as formal representation, querying, and reasoning.

\begin{figure}
    \centering
    \includegraphics[width=1\linewidth]{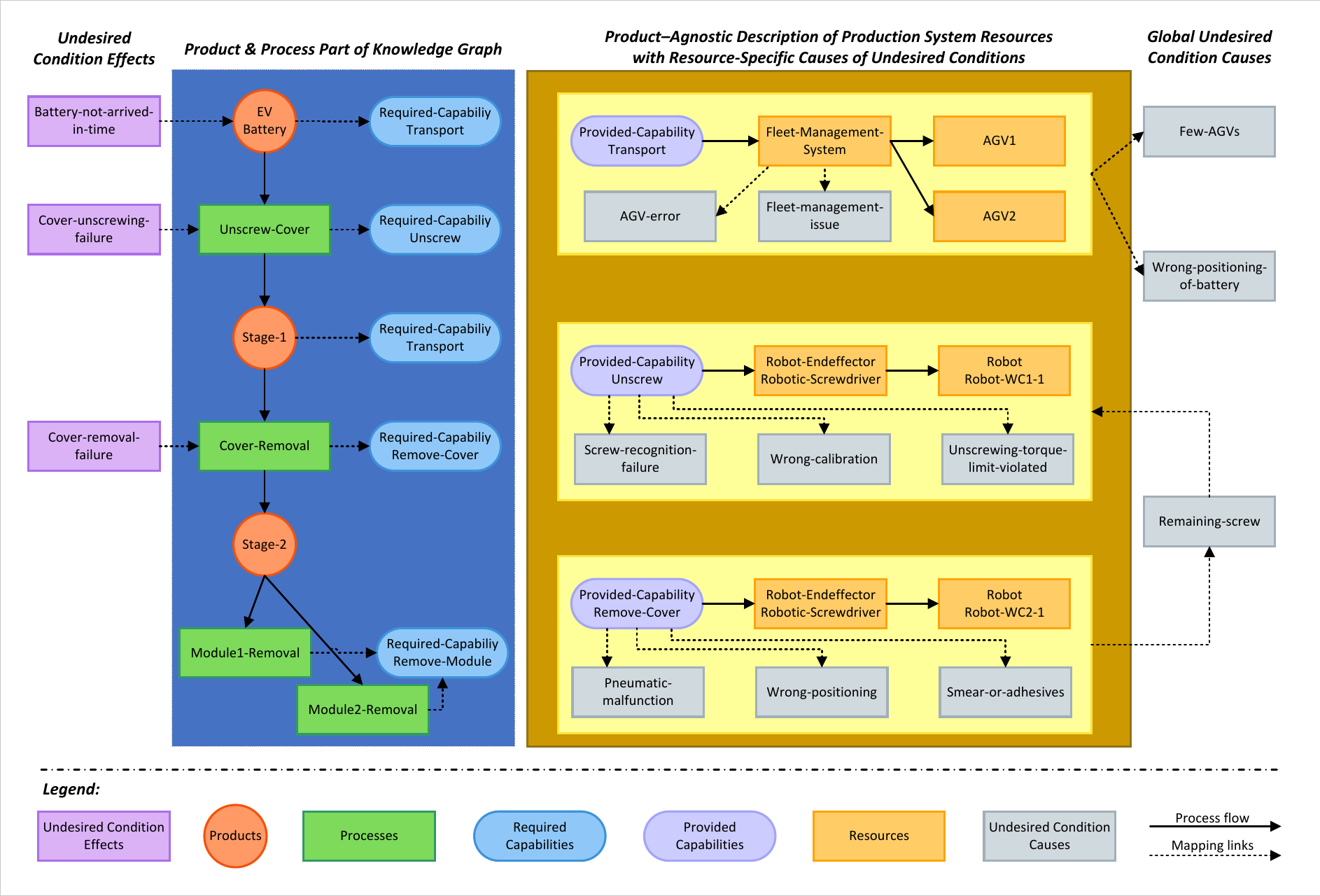}
    \caption{PPR-AKG: assets with required/provided capabilities, asset-specific/global undesired conditions.}
    \label{figModelDrivenProduction}
\end{figure}

Fig.~\ref{figModelDrivenProduction} illustrates the proposed comprehensive PPR-AKG,
which is structured into 4 columns. 
The core part of the knowledge graph is the blue region (i.e., second column), describing relationships between products (red circles) and processes (green boxes). 
To address Req.~1, processes are not assigned to resources, but processes are assigned to required capabilities (blue rounded boxes). 
Each required capability expresses a requirement for a resource, which allows to assign a resource flexibly.

Production resources, such as robotic workcells or autonomous vehicles, 
are specified in the yellow area by yellow boxes in Fig.~\ref{figModelDrivenProduction}. 
Resources provide one or more provided capabilities (purple rounded boxes). 
Whereas the products and processes in the blue region are described on the class level, meaning that $n$ instances of the product-and-process model can be instantiated and executed in parallel, production resources are described on the instance level. 
To execute production, the required capabilities have to be matched with the provided capabilities by solving a dynamic resource allocation 
task, which is supported by the PPR-AKG knowledge graph. 
An key benefit is the ability to add or remove capabilities on the fly, facilitating continuous engineering and improvement of the production systems.

For industrial practice and to meet Req.~2, it is very important to focus on the undesired conditions (purple boxes) as well as their plausible causes (gray boxes). 
Plausible causes can be expressed resource-specifically (i.e., inside yellow boxes, which can be defined by the resource vendor), as well as on the system level globally (right-hand column in Fig.~\ref{figModelDrivenProduction}). 
Such a scoping of causes for undesired conditions has been recommended by practitioners and it is also inspired by procan.do method~\cite{biffl2025introduction}.


While the knowledge graph provides explicit computer-understandable specification, it is difficult to interpret or to modify by humans. 
This gap can be efficiently bridged by Large Language Models (LLMs) or LLM-based chatbots. 
The primary use of LLMs to satisfy Req.~3 is encapsulating the PPR-AKG with an intuitive natural language interface for factory operators, production planners, maintenance technicians, and engineers. 
The typical prompts found a feasible schedule by matchmaking/scheduling required and provided capabilities and skills, or identified plausible causal explanations of undesired conditions (e.g., the prompt ``Why did the battery not arrive in time'' pinpointed possible causes to technicians). 
The secondary LLM use instantiates ontology individuals according to available production resources, but this topic required a cross-validation by judgment with different LLM types/versions.


\section{Conclusions and Future Work}

This paper reports on demonstrating the potential of semantic knowledge graphs for modeling and improving flexible Industry 4.0 production combined with analysis of undesired conditions in industrial production with plausible causes. 
The description logic in the proposed PPR-AKG ontology provides machine-understandable model of relevant knowledge, and the access to this model via LLMs facilitates natural human-oriented interaction with operators and engineers. 
The electric vehicle battery remanufacturing validation confirms practical applicability and substantial performance improvements.

While the PPR-AKG was found useful for mitigating undesired conditions in flexible production, the effective and efficient validation of the PPR-AKG content with LLMs remains a challenge.

In future work, we plan to systematically address the problem of instantiating ontology individuals with one LLM and judgment with another LLM. 
In addition, we plan to design an optimization method taking into account cost functions for undesired and desired conditions.


\begin{acknowledgments}
This work was co-funded by the European Union under the project ROBOPROX (reg. no. CZ.02.01.01/00/22\_008/0004590) and by the Rockwell Automation Laboratory for Distributed Intelligent Control (RA-DIC).
\end{acknowledgments}

\section*{Declaration on Generative AI}
  The author(s) have not employed any Generative AI tools.

\bibliography{references}




\end{document}